\documentclass[letterpaper, 10 pt, conference]{ieeeconf}  % Comment this line out if you need a4paper

\IEEEoverridecommandlockouts                              % This command is only needed if 
\usepackage{cite}
\usepackage{amsmath,amssymb,amsfonts}
\usepackage{algorithmic}
\usepackage{booktabs}
\usepackage{graphicx}
\usepackage{textcomp}
\usepackage{xcolor}
\usepackage{etoolbox}
\usepackage{graphicx}
\usepackage[skip=0.333\baselineskip]{caption}
\usepackage{subcaption}
\usepackage{multirow}
\usepackage{multicol}
\usepackage{tabularx}
\usepackage{hyperref}
\usepackage{romanbar}

\title{\LARGE \bf
VDT-Auto: End-to-end Autonomous Driving with VLM-Guided Diffusion Transformers}

\author{
    Ziang Guo \and 
    Konstantin Gubernatorov \and
    Selamawit Asfaw \and
    Zakhar Yagudin \and
    Dzmitry Tsetserukou
    \thanks{The authors are with the Intelligent Space Robotics Laboratory, Center for Digital Engineering, Skolkovo Institute of Science and Technology, Moscow, Russia
    \tt \{ziang.guo, Konstantin.Gubernatorov, Selamawit.Asfaw, Zakhar.Yagudin, d.tsetserukou\}@skoltech.ru}
}

\begin{document}

\maketitle
\thispagestyle{empty}
\pagestyle{empty}

%%%%%%%%%%%%%%%%%%%%%%%%%%%%%%%%%%%%%%%%%%%%%%%%%%%%%%%%%%%%%%%%%%%%%%%%%%%%%
\begin{abstract}
In autonomous driving, dynamic environment and corner cases pose significant challenges to the robustness of ego vehicle's decision-making. To address these challenges, commencing with the representation of state-action mapping in the end-to-end autonomous driving paradigm, we introduce a novel pipeline, VDT-Auto. Leveraging the advancement of the state understanding of Visual Language Model (VLM) with diffusion Transformer-based action generation, our VDT-Auto parses the environment geometrically and contextually for the conditioning of the diffusion process. Geometrically, we use a bird’s-eye view (BEV) encoder to extract feature grids from the surrounding images. Contextually, the structured output of our fine-tuned VLM is processed into textual embeddings and noisy paths. During our diffusion process, the added noise for the forward process is sampled from the noisy path output of the fine-tuned VLM, while the extracted BEV feature grids and embedded texts condition the reverse process of our diffusion Transformers. Our VDT-Auto achieved $0.52$ m on average L2 errors and $21\%$ on average collision rate in the nuScenes open-loop planning evaluation, presenting state-of-the-art performance. Moreover, the real-world demonstration exhibited prominent generalizability of our VDT-Auto. The code and dataset will be released at \href{https://github.com/ZionGo6/VDT-Auto}{\textit{https://github.com/ZionGo6/VDT-Auto}}.
\end{abstract}

\section{Introduction}

\subsection{Motivation}

Over time, diffusion model-based approaches have proven their value in robotic policy learning tasks \cite{chi2023diffusionpolicy, yang2024diff-es, yu2024ldp}. Dating back to the advancement of diffusion models, they have gained recognition as a cornerstone in the field of generative modeling \cite{peebles2023DiT}. Conditioned diffusion models extend vanilla diffusion models by incorporating additional information during the generation process, while latent diffusion models improve computational efficiency and sample quality by operating in a compressed latent space \cite{rombach2022latentDiff}. As shown above, diffusion models have exhibited promising potential in generating high-quality data across various modalities and improving the representation of complex data structures \cite{yang2023diffrep}. \par In robotic applications, multisensory data often includes rich and heterogeneous sources, such as camera images, LiDAR point clouds, etc. Diffusion models, through their ability to condition on various modalities, can generate coherent and contextually relevant outputs. This capability is particularly advantageous for robotic state-action mapping, where accurate interpretation and synthesis of multisensory inputs are crucial for effective decision-making and action execution \cite{li2025grmg}. \par Regarding autonomous driving, where the end-to-end paradigm has evolved vigorously \cite{sun2024sparsedrive}, state-action mapping is a core principle that enables vehicles to learn effective decision-making policies directly from raw sensor inputs \cite{liao2024diffusiondrive}. This process involves mapping the current state of the vehicle and its environment to appropriate control and planning actions. \par To enrich the state-understanding capacity of end-to-end autonomous driving systems, Visual Language Models (VLMs) have exhibited an outstanding impact, significantly improving the systems' interpreting capability of complex driving scenarios \cite{tian2024drivevlm, guo2024vlm-auto}. Accordingly, it is essential to enhance the decision-making capabilities as the improvement of state understanding by proposing adaptive and context-aware actions tailored to various driving scenarios \cite{li2024hydra}. 

\begin{figure}[t]
    \centering
    \includegraphics[width=0.98\linewidth]{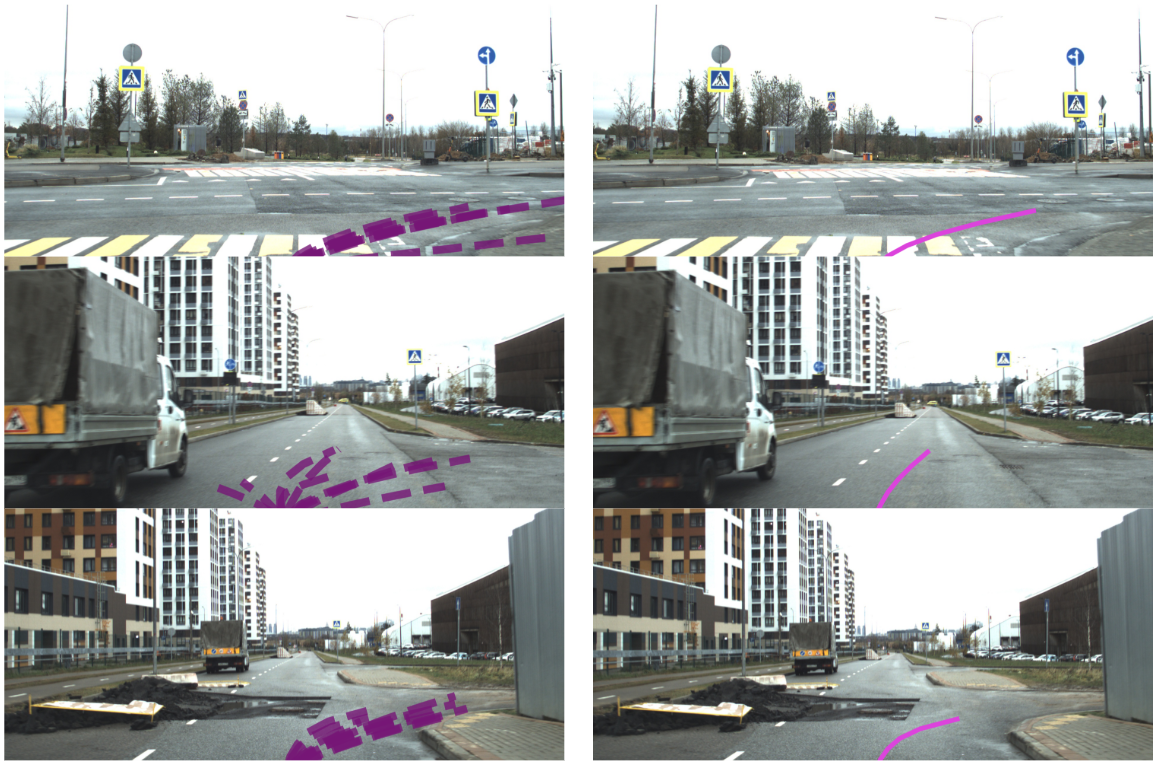}
    \begin{subfigure}{0.48\textwidth}
        \caption{VLM's path proposals from  \ \  (b) Conditional sampled paths \\ the continuous frames based on  \ \ by our diffusion Transformers. \\ a consistent scenario.}
        \label{fig:vlm_path_proposals}
    \end{subfigure}
    
    \caption{We conduct the experiments with our VDT-Auto on unseen real-world driving dataset in a zero-shot way. The VLM is fine-tuned on our processed nuScenes dataset while the path proposals from the fine-tuned VLM are able to provide contextual approximation across the unseen continuous frames. Subsequently, our diffusion Transformers sample the path proposals based on the geometric and contextual conditions.}
    \label{fig:cover}
\vspace{-0.5cm}
\end{figure}

\par With these insights, we propose VDT-Auto, an end-to-end paradigm that bridges states and actions via VLM and diffusion Transformers. For state understanding, images from the surrounding cameras are encoded into bird’s-eye view (BEV) features. In addition, a front image among the surrounding images is passed to a supervised fine-tuned VLM for contextual interpretation by the description of the detection, the advice of ego vehicle's behavior and the proposal of a path. Meanwhile, the designed diffusion Transformers encode both the BEV features from the BEV backbone and the contextual embeddings from the VLM as the states to predict the optimized path, where the added noise in the diffusion process is sampled from the VLM's proposed path. Our contributions in this paper are summarized as follows: 
\begin{itemize}
    \item We introduce a novel pipeline, VDT-Auto, which employs a BEV encoder and a VLM to geometrically and contextually parse the environment. The parsed information is then used to condition the diffusion process of our diffusion Transformers to generate the optimized actions of the ego vehicle.
    \item VDT-Auto is differentiable, where we use a processed nuScenes dataset to train our BEV encoder for perception and fine-tune our VLM for conditioning the diffusion Transformers. The constructed and processed dataset will be publicly available. 
    \item In the nuScenes open-loop planning evaluation, our VDT-Auto achieved $0.52$ m on average L2 errors and $21\%$ on average collision rate. In our real-world driving dataset, VDT-Auto showed promising performance on the unseen data in a zero-shot way.
\end{itemize}

\subsection{Related Work}

\subsubsection{Conditioned Diffusion Models}

By operating the data in latent space instead of pixel space, conditioned diffusion models have gained promising development \cite{rombach2022latentDiff}. MM-Diffusion \cite{ruan2023mmdi} designed for joint audio and video generation took advantage of coupled denoising autoencoders to generate aligned audio-video pairs from Gaussian noise. Extending the scalability of diffusion models, diffusion Transformers treat all inputs, including time, conditions, and noisy image patches, as tokens, leveraging the Transformer architecture to process these inputs \cite{bao2023ViTDiff}. In DiT \cite{peebles2023DiT}, William et al. emphasized the potential for diffusion models to benefit from Transformer architectures, where conditions were tokenized along with image tokens to achieve in-context conditioning. 

\subsubsection{Diffusion Models in Robotics}

Recently, a probabilistic multimodal action representation was proposed by Cheng Chi et al. \cite{chi2023diffusionpolicy}, where the robot action generation is considered as a conditional diffusion denoising process. Leveraging the diffusion policy, Ze et al. \cite{ze20243d} conditioned the diffusion policy on compact 3D representations and robot poses to generate coherent action sequences. Furthermore, GR-MG combined a progress-guided goal image generation model with a multimodal goal-conditioned policy, enabling the robot to predict actions based on both text instructions and generated goal images \cite{li2025grmg}. BESO used score-based diffusion models to learn goal-conditioned policies from large, uncurated datasets without rewards. Score-based diffusion models progressively add noise to the data and then reverse this process to generate new samples, making them suitable for capturing the multimodal nature of play data \cite{reuss2023md}. RDT-1B employed a scalable Transformer backbone combined with diffusion models to capture the complexity and multimodality of bimanual actions, leveraging diffusion models as a foundation model to effectively represent the multimodality inherent in bimanual manipulation tasks \cite{liu2024rdt-1b}. NoMaD exploited the diffusion model to handle both goal-directed navigation and task-agnostic exploration in unfamiliar environments, using goal masking to condition the policy on an optional goal image, allowing the model to dynamically switch between exploratory and goal-oriented behaviors \cite{sridhar2023nomad}. The aforementioned insights grounded the significant advancements of diffusion models in robotic tasks.

\subsubsection{VLM-based Autonomous Driving}

End-to-end autonomous driving introduces policy learning from sensor data input, resulting in a data-driven motion planning paradigm \cite{chen2024vadv2}. As part of the development of VLMs, they have shown significant promise in unifying multimodal data for specific downstream tasks, notably improving end-to-end autonomous driving systems\cite{ma2024dolphins}. DriveMM can process single images, multiview images, single videos, and multiview videos, and perform tasks such as object detection, motion prediction, and decision making, handling multiple tasks and data types in autonomous driving \cite{huang2024drivemm}. HE-Drive aims to create a human-like driving experience by generating trajectories that are both temporally consistent and comfortable. It integrates a sparse perception module, a diffusion-based motion planner, and a trajectory scorer guided by a Vision Language Model to achieve this goal \cite{wang2024hedrive}. Based on current perspectives, a differentiable end-to-end autonomous driving paradigm that directly leverages the capabilities of VLM and a multimodal action representation should be developed.

\section{Framework Overview}

\begin{figure*}
    \centering
    \includegraphics[width=0.95\linewidth]{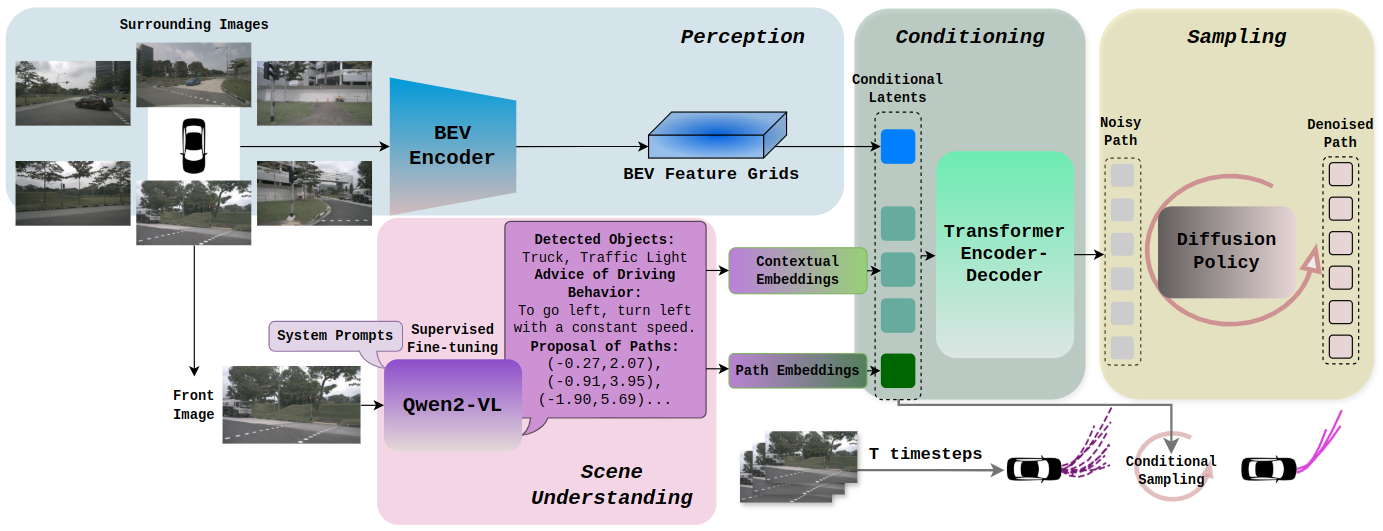}
    \caption{\textbf{Framework overview of VDT-Auto.} At each time step, the surrounding images are encoded by the BEV encoder to provide the geometric feature grids of the scenario. A front image from the surrounding images is analyzed by our fine-tuned VLM to provide the contextual information of the conditions. Based on the BEV feature grids and VLM output, we construct the conditional latents for our diffusion Transformers, where the BEV feature grids and VLM's detection and advice are embedded and VLM's path proposal is sampled for conditioning. In Section \Romannum{3}, we introduce our noise sampling approach in details. Finally, our diffusion Transformers denoise the VLM's path proposal, conditioning on the geometric feature grids of the scenario and the contextual information from our fine-tuned VLM.}
    \label{fig:overview}
\vspace{-0.3cm}
\end{figure*}

\subsection{BEV Encoder}

In Fig. \ref{fig:overview}, our BEV encoder is based on LSS \cite{philion2020lss, hu2021fiery}, where the surrounding camera images from the $T$ time steps are lifted into the BEV feature grids. $F^{k}_{t}\in\mathbb{R}^{(C_{f} + D_{d}) \times H \times W}$ represents the extracted features of the $k$-th camera at time $t$ from the image backbone, where $F^{k}_{t,C_{f}} \in \mathbb{R}^{C_{f} \times H \times W}$ is the contextual features and $F^{k}_{t,D_{d}} \in \mathbb{R}^{D_{d} \times H \times W}$ represents the estimated depth distribution. Then the contextual feature map in height dimension $F'^{k}_{t}$ is computed as $F^{k}_{t,C_{f}} \otimes F^{k}_{t,D_{d}}$. According to the nuScenes camera setup \cite{caesar2020nuscenes}, with the intrinsics and extrinsics of the cameras, $F'^{k}_{t}$ is then aggregated and weighted along the height dimension into the ego-centered coordinate system to obtain the BEV feature grids $G_{t} \in \mathbb{R}^{C_\text{state} \times H \times W}$ at time $t$, where $C_\text{state}$ is the number of state channels.

\subsection{VLM Module}

For our work, Qwen2-VL-7B is used to bridge the input of sensory data and the output of contextual conditions \cite{wang2024qwen2vl}. In Qwen2-VL, Multimodal Rotary Position Embedding (M-RoPE) is applied to process multimodal input by decomposing rotary embedding into temporal, height, and width components, which equips Qwen2-VL with powerful multimodal data handling capabilities. In our VDT-Auto, the supervised fine-tuning of Qwen2-VL-7B is carried out by feeding a front image of surrounding cameras and system prompts, expecting the output of the description of the detection, the structured advice of the behavior of the ego vehicle, and the proposal of a path. To achieve supervised fine-tuning, we constructed our fine-tuning dataset by extracting ground truth information from the nuScenes dataset \cite{caesar2020nuscenes}. In Section \Romannum{3}, we will introduce more details about our dataset construction and supervised fine-tuning.

\subsection{Diffusion Prerequisites}

In Fig. \ref{fig:overview}, we show our entire VDT-Auto pipeline, where the feature grids $G_{t} \in \mathbb{R}^{C_\text{state} \times H \times W}$ of the BEV encoder and the contextual output $S_t$ of Qwen2-VL-7B including the description of the detection and structured advice on the behavior of the ego vehicle are encoded as state conditions for the diffusion process. Thus, in our designed diffusion Transformers, the conditioned policy $\pi_{\theta}(A_t | G_t, S_t)$ predicts the denoised path $A_t = (a^0_t, a^1_t, \ldots,a^n_t)$ of length $n$, conditioned on both the current BEV features $G_t$ and contextual embeddings $S_t$ \cite{bao2023onemmdi, han2024emma, reuss2024mdt}. During training, the proposal of a path based on supervised fine-tuning of Qwen2-VL-7B at time $t$ pairing with current BEV features $G_t$ and contextual embeddings $S_t$ to form the training set, where our diffusion Transformers aim to maximize log-likelihood $\ell_{\text{training}}$ throughout the training set,

\vspace{-0.3cm}
\begin{equation}
    \ell_{\text{training}} = \underset{\theta}{\arg\max}{}_{(a_t^i, g_t^i, s_t^i) \in (A'_t, G_t, S_t)} \log {\pi_\theta}({a_t^i | g_t^i, s_t^i}),
\end{equation}
where $a_t^i, g_t^i, s_t^i$ are sampled from our constructed training set. We extract the noise distribution $\sigma_{\text{VLM}}$ from the path output of the supervised fine-tuned Qwen2-VL-7B to construct the noisy path dataset $A'_t$ by adding the sampled noise from the extracted noise distribution $\sigma_{\text{VLM}}$ to the ground truth path $A_{gt}$ of nuScenes.

\par In Section \Romannum{3}, we demonstrate that the noise distribution $\sigma_{\text{VLM}}$ of the path proposal from our fine-tuned Qwen2-VL-7B is treated as a normal distribution, where we examine the extracted noise using One-Sample Kolmogorov-Smirnov test for both the $x$ and $y$ coordinates of the paths \cite{Kstest}. 

\subsection{Loss Functions}

Our diffusion Transformers predict the denoised path $A_t$ conditioned on current BEV features $G_t$ and contextual embeddings $S_t$. Therefore, the loss function is defined as follows.

\vspace{-0.3cm}
\begin{equation}
\begin{split}
    \mathcal{L}_{\text{train}} = \mathcal{L}_{\text{MSE}} (\pi_{\theta}(A_t | g_t, s_t, \boldsymbol{\epsilon}), A_{gt}) + \\
    \mathcal{L}_{\text{MSE}}(\sum_{j=1}^{n} a^j, \sum_{j=1}^{n} a_{gt}^j),
\end{split}
\end{equation}
where $\pi_{\theta}$ is our trained diffusion Transformers. Under the conditions of encoded BEV features $g_t \in G_t$, contextual embeddings $s_t \in S_t$, and added noise $\boldsymbol{\epsilon} \in \sigma_{\text{VLM}}$, the first part of our loss function is the mean squared error between the path prediction $A_t$ and the ground truth path from nuScenes $A_{gt}$. Besides, the second part of our loss function is the mean squared error between the cumulative sum of the waypoints $a^j \in A_t$ and $a_{gt}^j \in A_{gt}$.

\section{Methodology}

\begin{figure}
    \centering
    \includegraphics[width=0.8\linewidth]{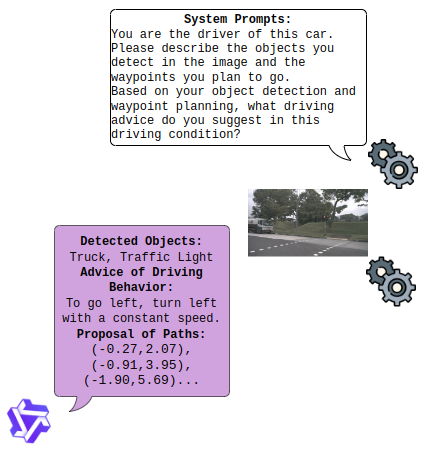}
    \caption{In our constructed dataset for VLM's supervised fine-tuning, we input a system prompt and a front image into VLM and expect a structured output including detected objects, advice of the ego vehicle's driving behavior, and the proposal of a path.}
    \label{fig:sft_dataset}
\vspace{-0.3cm}
\end{figure}

\subsection{Supervised Fine-tuning of Qwen2-VL-7B}

We extract the information including the detection results and ego future trajectory from nuScenes. Then we generate the advice for the ego vehicle's behavior according to the changes in the ego vehicle's speed and temporal trajectory. By feeding the system prompts and a front image from the surrounding cameras into the VLM, an example of our fine-tuning dataset is shown in Fig. \ref{fig:sft_dataset}.

\begin{table}[ht]
    \caption{The results of the verification of normally distributed noise from our fine-tuned VLM's responses.}
    \centering
    \begin{tabular}{@{}cccll@{}}
\cmidrule(r){1-3}
\multicolumn{1}{c|}{\begin{tabular}[c]{@{}c@{}}Number of VLM \\ path outputs\end{tabular}} & \multicolumn{1}{c|}{\begin{tabular}[c]{@{}c@{}}Number of the paths \\ with normally \\ distributed noise\end{tabular}} & \begin{tabular}[c]{@{}c@{}}Percentage of the paths \\ with normally \\ distributed noise\end{tabular} &  &  \\ \cmidrule(r){1-3} 
\multicolumn{1}{c|}{20235} & \multicolumn{1}{c|}{19830} & 98.00\% &  &  \\ \cmidrule(r){1-3}
\multicolumn{1}{c|}{37812} & \multicolumn{1}{c|}{36890} & 97.60\% &  &  \\ \cmidrule(r){1-3}
\multicolumn{1}{c|}{67113} & \multicolumn{1}{c|}{65453} & 97.52\% &  &  \\ \cmidrule(r){1-3}
\multicolumn{1}{c|}{111384} & \multicolumn{1}{c|}{108497} & 97.40\% &  &  \\ \cmidrule(r){1-3} 
&  &                &  & 
\end{tabular}
    \label{tab:vlm_noise}
\vspace{-1cm}
\end{table}

\subsection{VLM-guided Diffusion Transformers}

\textbf{Normal distribution verification.} To obtain the responses from our VLM module, we iteratively feed the system prompts and the front image of the surrounding cameras per time $t$ to our fine-tuned VLM throughout the training set to perform inference. Based on all the responses obtained, we extract the noise distribution $\sigma_{\text{VLM}}$ from the coordinates $x$ and $y$ of the proposal of the paths by subtracting the corresponding ground truth paths $A_{gt}$. In Table \ref{tab:vlm_noise}, we show that the percentage of paths with normally distributed noise according to the amounts of the proposal of the paths obtained, where the One-Sample Kolmogorov-Smirnov test is used on both $x$ and $y$ coordinates to compare the empirical cumulative distribution function (EDF) of our noise data against the theoretical cumulative distribution function (CDF) of a normal distribution with the same mean and standard deviation as our noise data due to the nonparametric attributes of the Kolmogorov-Smirnov test \cite{Kstest}. Given a path sample $a^0, a^1, \ldots, a^n$, the empirical distribution function (EDF) $F_n(x)$ is defined as

\begin{equation}
    F_n(x) = \frac{1}{n} \sum_{i=1}^{n} I(a^i \leq x),
\end{equation}
where $I(\cdot)$ is the indicator function that equals 1 if the condition inside is true and 0 otherwise.

The Kolmogorov-Smirnov distribution $D_n$ is defined as the maximum absolute difference between the EDF $F_n(x)$ and the CDF $F(x)$,

\begin{equation}
    D_n = \sup_x |F_n(x) - F(x)|,
\end{equation}
where $\sup$ is the supremum of the set of distances. 

For our path sample $a^0, a^1, \ldots, a^n$, $p$ is defined as the probability that the Kolmogorov distribution $K$ exceeds the calculated $D_n$. If $p$ is below the significance level $\alpha_p$, the path sample does not follow a normal distribution. Setting $\alpha_p = 0.05$, the Kolmogorov distribution $K$ is defined by its cumulative distribution function,

\begin{equation}
    P(K \leq t) = 1 - 2 \sum_{k=1}^{\infty} (-1)^{k-1} e^{-2 k^2 t^2},
\end{equation}
for $t > 0$.

Thus, when $p = \Pr(K > D_n) < \alpha_p$, the noise is considered as being drawn from a normal distribution for both the coordinates $x$ and $y$ of the proposal of the paths. During the verification, we iteratively input the front images from the nuScenes dataset into our fine-tuned VLM to obtain the responses. We then identify the paths with normally distributed noise from these responses, where the normal distribution is verified via One-Sample Kolmogorov-Smirnov test for both the $x$ and $y$ coordinates of the paths. In Table \ref{tab:vlm_noise}, we show the number and percentage of the paths qualified by the Kolmogorov-Smirnov test on both $x$ and $y$ coordinates across different samples from the responses obtained.

\textbf{Diffusion process.} Based on the DDIM scheduler \cite{song2020ddim}, we gradually add the sampled noise from the VLM to the ground truth nuScenes path $A_{gt}$ in a forward diffusion process as follows, 
\begin{equation}
    a^{i'} = \sqrt{\bar{\alpha}^i}{a}_{gt}^0 + \sqrt{1-\bar{\alpha}^i}\boldsymbol{\epsilon}, \quad \boldsymbol{\epsilon} \sim \sigma_{\text{VLM}},
\end{equation}
where $a^{i'}$ is the noised sample at scheduler timestep $i$. $\boldsymbol{\epsilon}$ is the sampled noise from $\sigma_{\text{VLM}}$. $\beta^t$ is the sequence of variances that control the amount of noise added at each diffusion timestep and $\bar{\alpha^i}=\prod_{t=1}^i\alpha^t=\prod_{t=1}^i(1-\beta^t)$. To ensure the stability of our training and unbiased noising, we then standardize the noised path $a^{i'} \in {A'_t}$.

In the reverse diffusion process, our diffusion Transformers $\pi_\theta$ predict the denoised path $A_t$ for each denoising timestep $t_\text{R}$, where the noisy path from the output of our VLM is denoised following the DDIM scheduler. We define the noise scheduler as $\sigma(t_\text{R}) = e^{-t_\text{R}}$. Given the input of the noisy path $A'_t$ and the prediction $A_t$, the denoised path is updated as follows,

\begin{equation}
    A'_t = \left( \frac{\sigma(t_\text{R+1})}{\sigma(t_\text{R})} \right) \cdot A'_t - (e^{-h}-1) \cdot {A_t},
\end{equation}
where $h = t_\text{R+1} - t_\text{R}$ is the denoising time interval.

\begin{table}[]
\caption{Ablation study of our timestep embedding (TSE), cross-attention-based fusion of geometric and contextual embedding (CAF), contextual average pooling (CAP), and BEV feature compression (BFC) on nuScenes validation set.}
\centering
\small{
\begin{tabular}{cccccc}
\hline
\toprule
TSE & CAF & CAP & \multicolumn{1}{c|}{BFC} & \multicolumn{1}{c|}{Avg. L2} & Avg. Collision Rate \\ \hline
\midrule
 \checkmark & \checkmark & \checkmark & \multicolumn{1}{c|}{\checkmark}  & \multicolumn{1}{c|}{0.52}       &    0.21    \\
  & \checkmark & \checkmark & \multicolumn{1}{c|}{\checkmark}  & \multicolumn{1}{c|}{1.08}       &    0.60    \\
 \checkmark & \checkmark &   & \multicolumn{1}{c|}{}  & \multicolumn{1}{c|}{1.21}       &    0.88    \\ \hline
  &  &   &                        &                             &       

\end{tabular}
}
\label{tab:ablation}
\vspace{-0.5cm}
\end{table}

\begin{figure}
    \centering
    \includegraphics[width=0.88\linewidth]{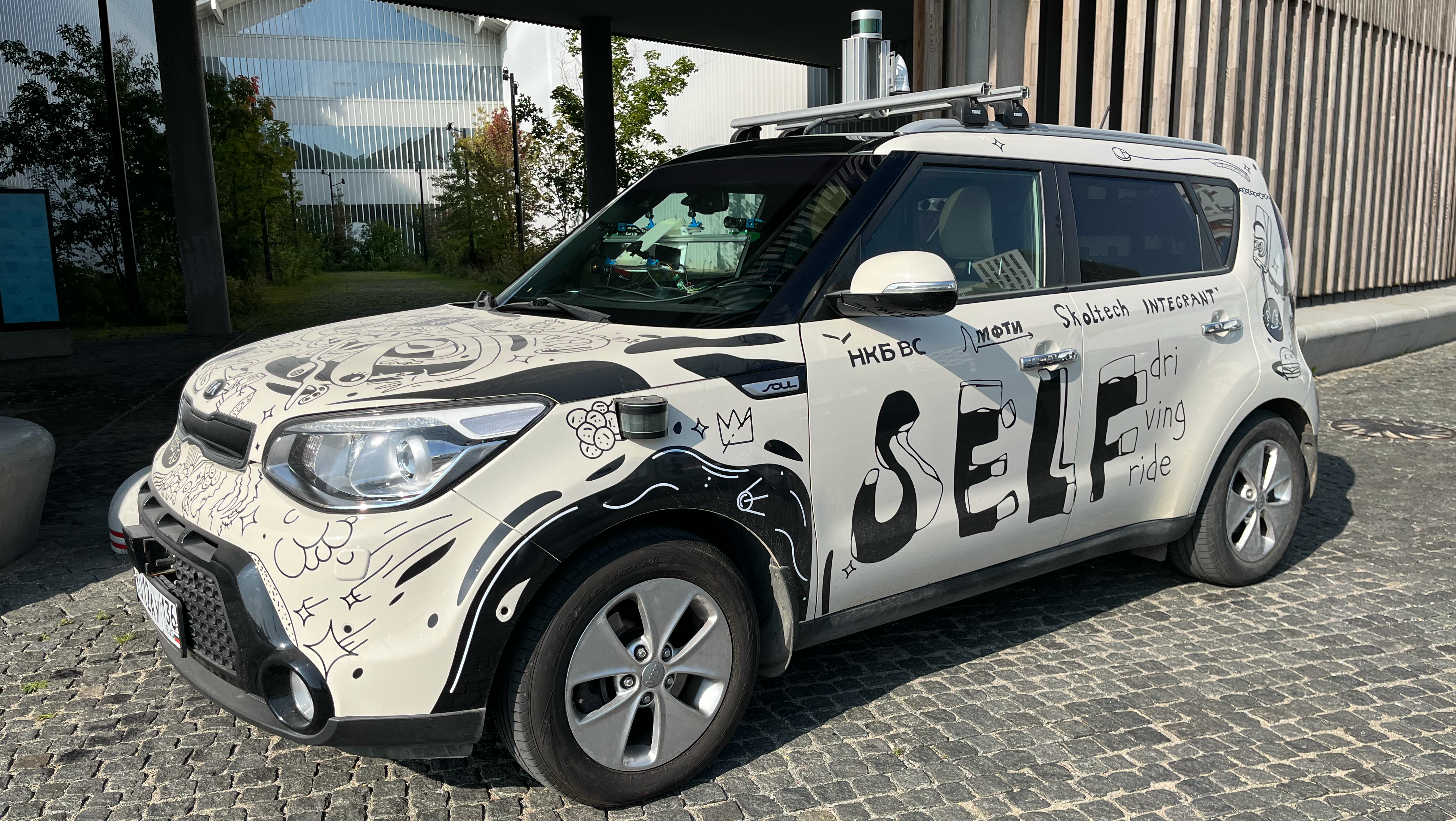}
    \caption{Our experimental car for the recording of our real-world driving dataset.}
    \label{fig:sdc}
\vspace{-0.5cm}
\end{figure}

\section{Experiments}

\begin{table*}[]
\caption{The open-loop planning results of our VDT-Auto on nuScenes validation set.}
\centering
\resizebox{0.66\linewidth}{!}{
    \begin{tabular}{l|l|c|c|c|c|c|c|c|ccc}
        \toprule
        \multirow{2}{*}{No.} & \multirow{2}{*}{Methods} & \multicolumn{4}{c|}{L2 (m) $\downarrow$} & \multicolumn{4}{c}{Collision Rate (\%) $\downarrow$} \\
         & & 1s & 2s & 3s & Avg. & 1s & 2s & 3s & Avg. \\
        \midrule
        1 & FF~\cite{hu2021safeff} & 0.55 & 1.20 & 2.54 & 1.43 & 0.06 & 0.17 & 1.07 & 0.43 \\
        2 & EO~\cite{kh2022differentiable} & 0.67 & 1.36 & 2.78 & 1.60 & 0.04 & 0.09 & 0.88 & 0.33 \\
        3 & ST-P3~\cite{hu2022st-p3} & 1.33 & 2.11 & 2.90 & 2.11 & 0.23 & 0.62 & 1.27 & 0.71 \\
        4 & UniAD~\cite{hu2023uniAD} & 0.48 & 0.96 & 1.65 & 1.03 & 0.05 & 0.17 & 0.71 & 0.31 \\
        5 & GPT-Driver~\cite{mao2023gpt} & 0.27 & 0.74 & 1.52 & 0.84 & 0.07 & 0.15 & 1.10 & 0.44 \\
        6 & VLP-UniAD~\cite{pan2024vlp} & 0.36 & 0.68 & 1.19 & 0.74 & 0.03 & 0.12 & 0.32 & 0.16 \\
        7 & RDA-Driver~\cite{huang2024rda} & 0.23 & 0.73 & 1.54 & 0.80 & \textbf{0.00} & 0.13 & 0.83 & 0.32 \\
        8 & DriveVLM~\cite{tian2024drivevlm} & \textbf{0.18} & \textbf{0.34} & \textbf{0.68} & \textbf{0.40} & 0.10 & 0.22 & 0.45 & 0.27 \\
        9 & HE-Drive-B~\cite{wang2024hedrive} & 0.30 & 0.56 & 0.89 & 0.58 & \textbf{0.00} & \textbf{0.03} & \textbf{0.14} & \textbf{0.06} \\
        
        \midrule
        10 & \textbf{Ours} & 0.20 & 0.47 & 0.88 & 0.52 & 0.05 & 0.18 & 0.40 & 0.21 \\

        \midrule
       
    \end{tabular}
}
\label{tab:L2}
\vspace{-0.3cm}
\end{table*}

\subsection{Experimental Setup}

We conducted our training and open-loop experiments on the nuScenes dataset that consists of $1,000$ street scenes collected from Boston and Singapore, known for their dense traffic and challenging driving conditions \cite{caesar2020nuscenes}. In addition, we evaluated our pipeline on a real-world driving dataset in a zero-shot manner. The real-world driving dataset was recorded by our experimental car shown in Fig. \ref{fig:sdc} \cite{guo2024hawkdrive}. \par In our experiments, we first trained our BEV encoder and fine-tuned Qwen2-VL-7B on nuScenes to obtain the BEV features and the VLM responses from their inference. Then we cached the BEV features and VLM responses and constructed the training set for the training of our diffusion Transformers.

\subsection{Comparison with Other State-of-the-Art Methods on nuScenes}

During the evaluation, on the nuScenes validation set \cite{caesar2020nuscenes}, we compared our VDT-Auto with other methods by the L2 error in meters and the collision rate in percentage in Table \ref{tab:L2}. The average L2 error is determined by calculating the distance between each waypoint in the planned trajectory and the corresponding waypoint in the ground truth trajectory. This metric indicates how closely the planned trajectory aligns with a human-driven trajectory. The collision rate is assessed by positioning an ego-vehicle bounding box at each waypoint along the planned trajectory and subsequently checking for any intersections with the ground truth bounding boxes of other objects. Our VDT-Auto achieved state-of-the-art performance in open-loop planning tasks.

\subsection{Experiments on Real-world Driving Dataset}

In our real-world driving dataset, we demonstrate the potentials of our VDT-Auto on unseen data in a zero-shot way, where our BEV encoder is adjusted to obtain the extracted features from a single front image, and the fine-tuned VLM analyzes the front image to provide the contextual information. In Fig. \ref{fig:cover}, we show the VLM's path proposals from the continuous frames based on a consistent scenario in the left column (a), while the conditional sampled paths are shown in the right column (b).

\subsection{Ablation Study}

To verify the effectiveness of our design, in Table \ref{tab:ablation}, we show the results of the ablation experiments of our VDT-Auto with timestep embedding (TSE), cross-attention-based fusion of geometric and contextual embedding (CAF), contextual average pooling (CAP), and BEV feature compression (BFC) in nuScenes validation set. In TSE, we embed the noise scheduler timesteps for the preparation of a cross-attention-based fusion of geometric and contextual embedding, while CAP and BFC are the dimensionality reduction of the BEV features $G_t$ and contextual embeddings $S_t$ for the stability of training and inference.

\section{Conclusion}

Considering the advancements of state understanding and the corresponding decision-making capabilities, we propose a novel pipeline, VDT-Auto, where the state information is encoded geometrically and contextually, conditioning a diffusion Transformer-based action generation. In this paper, we demonstrate the methodology of using powerful VLM such as Qwen2-VL to bridge states and conditions, as well as the connections between conditions and actions via a diffusion policy. The verification of our VDT-Auto was performed using nuScenes open-loop planning evaluation, where our VDT-Auto achieved $0.52$ m on average L2 errors and $21\%$ on average collision rate. In addition, on a real-world driving dataset, our VDT-Auto shows its promising generalizability. \par During our development, we discovered that the distribution of training data had varying influences on the different parts of our VDT-Auto due to the model scales of the VLM and the diffusion model-based network. Therefore, with sufficient computational resources, an end-to-end training approach should be developed to mitigate the influence of data distribution. In our future work, with an end-to-end training pipeline, VDT-Auto will be targeting towards more complex traffic scenarios and a close-loop evaluation. Owing to the rapid evolution of VLMs and robotic policy, VDT-Auto is able to contribute as a cornerstone case in data-driven policy learning tasks.

% \section*{Acknowledgment}

%%%%%%%%%%%%%%%%%%%%%%%%%%%%%%%%%%%%%%%%%%%%%%%%%%%%%%%%%%%%%%%%%%%%%%%%%%%%%%%%%%%%%%%%%%%
\newpage
\bibliographystyle{IEEEtran}
\bibliography{references}

\end{document}